\documentclass[conference]{IEEEtran}
\IEEEoverridecommandlockouts    % to override locked commands

\usepackage{multirow}
\usepackage{lipsum}
\usepackage{amsmath}
\usepackage{amssymb}
\usepackage[T1]{fontenc}
\usepackage[ruled,vlined]{algorithm2e}
\usepackage{graphicx}
\graphicspath{{./Figures/}}
\usepackage{float}
\usepackage{booktabs}

\title{\LARGE \bf iPay: Integrated Payment Action Recognition via Multimodal Networks and Adaptive Spatial Prior Learning}

\author{
	\parbox{\textwidth}{%
		\centering
		Kaicong Huang$^{1}$, Weiheng Oh$^{1}$, Thomas Guggisberg$^{2}$, Ruimin Ke$^{1\star}$%
	}%
	\thanks{$^{1}$ Rensselaer Polytechnic Institute, 110 Eighth Street, Troy, NY USA 12180.
		{\tt\small huangk10@rpi.edu, ohw@rpi.edu, ker@rpi.edu}}%
	\thanks{$^{2}$ Capital District Transportation Authority, 110 Watervliet Avenue, Albany, NY USA 12206.
		{\tt\small thomas@cdta.org}}%
	\thanks{$^{\star}$ Corresponding author.}%
}

\begin{document}
	
	\maketitle
	\thispagestyle{empty}
	\pagestyle{empty}

	%%%%%%%%%%%%%%%%%%%%%%%%%%%%%%%%%%%%%%%%%%%%%%%%%%%%%%%%%%%%%%%%%%
\begin{abstract}
Automated transit payment analysis is vital for scalable fare auditing and passenger analytics, yet practice still relies on limited manual inspection. Prior vision- and skeleton-based methods remain brittle under noisy onboard surveillance and often depend on poorly generalizable handcrafted features.
Building on the success of graph convolutional networks in human action recognition, we observe that skeleton features excel at modeling global spatiotemporal dependencies but tend to underemphasize the subtle local relative motions that distinguish payment actions. In contrast, RGB features preserve fine-grained spatial details yet often lack reliable temporal continuity in surveillance footage. 
To bridge both system-level deployment needs and model-level design challenges, we present iPay, an integrated payment action recognition framework for onboard transit surveillance system. iPay adopts a multimodal mixture-of-experts architecture with four tightly coupled streams: (1) an RGB expert stream emphasizing local evidence via region-focused computation; (2) a skeleton expert stream modeling articulated motion with a graph convolutional backbone; (3) a dual-attention fusion stream enabling skeleton-to-RGB temporal transfer and RGB-to-skeleton spatial enhancement; and (4) a prior-driven Spatial Difference Discriminator (SDD) that explicitly models hand-to-anchor relative motion to improve task-specific discriminability.
We also collaborate with local transit agencies to collect over 55 hours of real onboard surveillance footage, yielding 500+ payment clips. Experiments show that iPay outperforms prior methods and achieves 83.45\% recognition accuracy with competitive computational efficiency, making it suitable for edge deployment. Code is available at https://github.com/ccoopq/iPay.
\end{abstract}

	%%%%%%%%%%%%%%%%%%%%%%%%%%%%%%%%%%%%%%%%%%%%%%%%%%%%%%%%%%%%%%%%%%
	\section{Introduction}
	Reliable fare collection is fundamental to public transit operations, affecting revenue stability, service planning, and equity policies. Accordingly, automated fare collection and payment analytics have become key components of modern intelligent transit systems, enabling large-scale passenger behavior analysis and system monitoring \cite{bieler2022survey}.
	In many proof-of-payment networks, manual fare inspection remains the primary method but is labor-intensive and hard to scale in dense, high-volume transit environments \cite{barabino2024fare}, highlighting the need for more cost-effective and scalable auditing mechanisms \cite{reddy2011measuring}.

	Recent advances in computer vision have made it feasible to consider automated recognition of passenger payment actions using existing onboard cameras, which are already deployed at scale for security and incident review. In principle, an integrated payment recognition system could provide continuous, non-intrusive monitoring for ridership analytics without installing additional hardware \cite{huang2025transitreid}. In practice, however, three persistent barriers hinder the deployment of such systems in real-world transit settings. First, training data are scarce because payment behavior is rarely captured and labeled at scale in the wild. Second, onboard surveillance footage has its domain-specific challenges, including low-resolution and noisy, with motion blur, vibration, reflections, and partial occlusions, which substantially degrades the performance of general-purpose video understanding models \cite{ciampi2022bus}. Third, privacy constraints discourage reliance on identifiable appearance cues, highlighting the need for privacy-aware representations and processing pipelines.

	\begin{figure}
		\centering
		\includegraphics[width=\columnwidth]{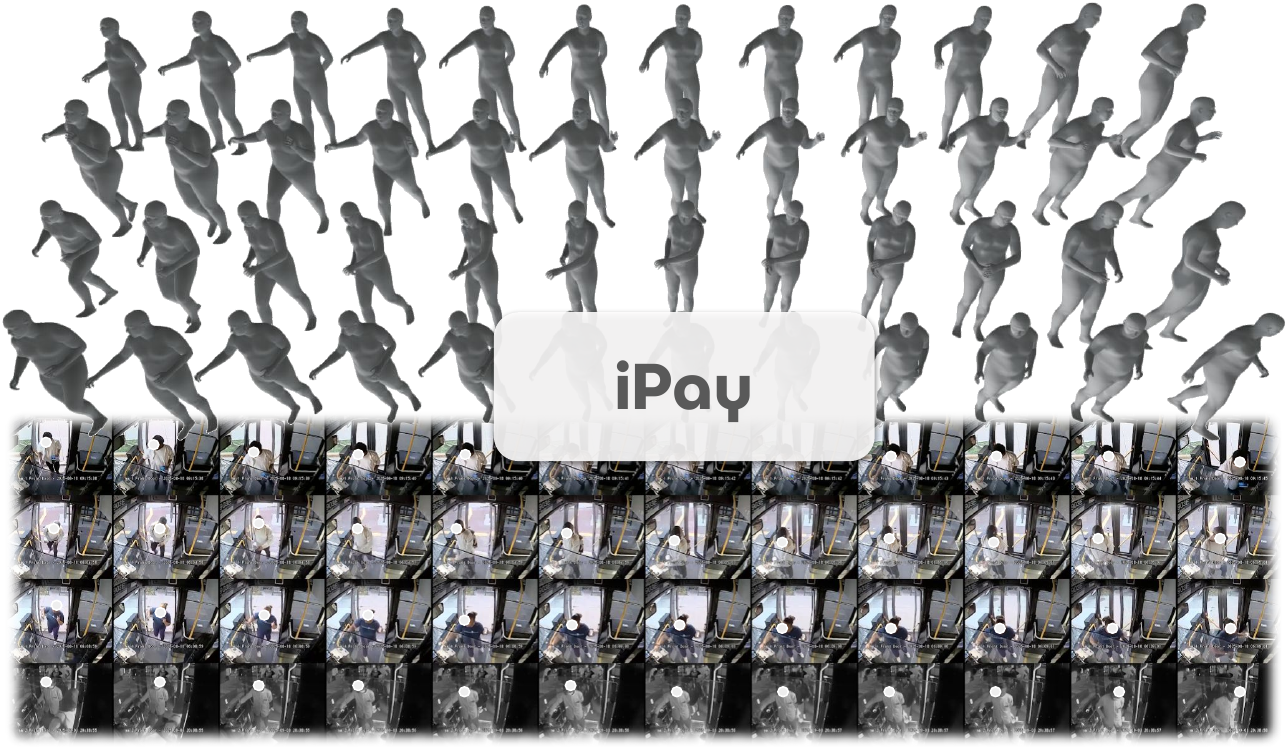}
		\caption{Qualitative examples from our collected onboard payment-action dataset, illustrating four payment modalities (Cash, QR Code, Tap, and Swipe). For each modality, we show the corresponding RGB frames and skeleton sequence extracted from the same clip, highlighting the fine-grained local hand motion patterns that motivate our multimodal design.}
		\label{fig:intro}
	\end{figure}

	From a modeling perspective, existing action recognition methods are primarily designed for generic video scenarios and do not explicitly encode the task structure of transit payment behaviors. Strong video backbones, including SlowFast \cite{feichtenhofer2019slowfast} and transformer-based architectures such as Video Swin \cite{liu2022video} and self-supervised pretraining frameworks such as VideoMAE \cite{tong2022videomae}, have demonstrated impressive performance on generic benchmarks, but their accuracy can drop sharply under surveillance-style conditions where the payment action occupies a small region and the scene contains heavy distractions.
	Skeleton-based recognition provides an appealing alternative because it abstracts away appearance and illumination, and graph-based formulations such as ST-GCN \cite{yan2018spatial} and its successors have become a dominant paradigm for spatiotemporal modeling of articulated motion \cite{shi2019two,chen2021channel,chi2022infogcn,zhou2024blockgcn,myung2024degcn}. Nonetheless, payment actions are inherently fine-grained and defined by highly localized hand-to-terminal relative kinematics (e.g., lateral swipes versus short contact-and-dwell during tapping), which can be overlooked when models emphasize whole-body motion or lack explicit task-oriented geometric modeling.

\begin{figure*}
		\centering
		\includegraphics[width=\textwidth]{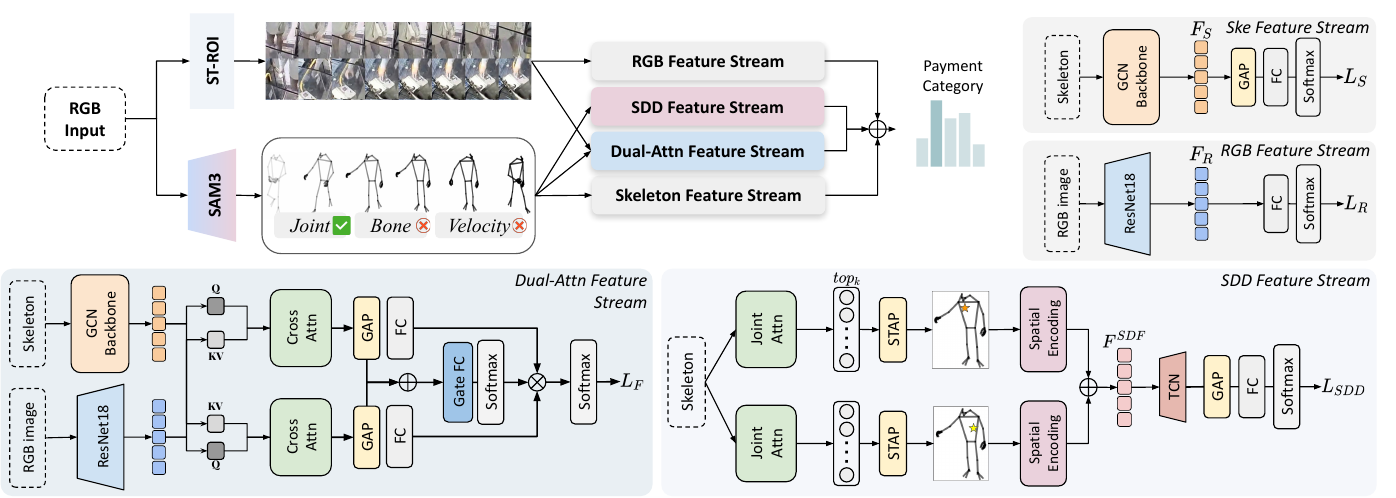}
		\caption{Overview of the proposed iPay framework. The system integrates onboard surveillance RGB and skeleton streams within a multimodal mixture-of-experts architecture consisting of an RGB expert, a skeleton expert, dual-attention fusion, and a Spatial Difference Discriminator (SDD). During training, all streams are jointly optimized with their losses summed into the final objective. During inference, the logits are aggregated to obtain the final prediction.}
		\label{fig:framework}
	\end{figure*}

	This paper presents \textbf{iPay}, an \textbf{i}ntegrated \textbf{pay}ment action recognition framework designed specifically for onboard transit surveillance. Fig. \ref{fig:framework} illustrates the overall framework. \textbf{At the system level}, iPay is designed to integrate into existing transit monitoring infrastructure, enabling passive sensing without deploying extra hardware. We also collaborate with a local transit authority to collect and curate over 55 hours of real onboard surveillance footage, yielding more than 500 passenger payment clips to form a payment action recognition dataset. \textbf{At the model level}, iPay integrates both RGB and skeleton modalities that compensate for each other. It leverages surveillance videos as the input and derives a privacy-aware RGB representation with locally cropped regions, and a skeletal representation using a modern foundation model pipeline. In particular, SAM3 3D body model \cite{carion2025sam} makes it practical to extract stable body-and-hand structure under challenging viewpoints and occlusions, which is critical for payment behaviors dominated by local hand motion.

	To address the domain gap and the task-specific nature of payment behaviors, iPay adopts a multimodal mixture-of-experts architecture with four tightly coupled streams. 
	(1) An RGB expert stream emphasizes local evidence by focusing computation on the most informative regions, reducing background distraction typical in surveillance footage. 
	(2) A skeleton expert stream models the spatiotemporal dependencies of articulated motion using a graph convolutional backbone and remains robust to varying visual quality.
	(3) A dual-attention fusion stream. Skeleton expert is effective at capturing global spatiotemporal dependencies but can under-emphasize the subtle relative local motions that are critical for payment actions, whereas RGB expert preserves fine-grained spatial details yet often lack reliable temporal continuity. To leverage these complementary strengths, we design a dual-attention interaction module in which each modality attends to and guides the other, enabling mutual refinement of temporal structure and spatial detail. 
	(4) A Spatial Difference Discriminator (SDD) stream injects payment-specific priors by explicitly modeling local hand motion relative to adaptively learned anchors, enabling the network to capture discriminative hand-to-anchor dynamics that are easily overlooked by generic backbones.
	
	Unlike prior ensemble approaches that only combine logits at the final stage \cite{chen2021channel, myung2024degcn, bruce2022mmnet}, our fusion is in feature-level, and the four streams are trained end-to-end. Notably, we use only joint coordinates as the skeleton representation, rather than incorporating bone or velocity features as in \cite{chen2021channel,myung2024degcn}, while still outperforming state-of-the-art methods on the payment-action recognition task. Another key design principle in iPay is privacy-aware processing. Instead of relying on full-frame identity cues, the method operates on cropped regions and skeletal data, providing a compact representation that naturally reduces appearance leakage.

	The main contributions are summarized as follows:
	\begin{itemize}
	\item We propose a deeply coupled multimodal mixture-of-experts architecture that is jointly optimized to capture fine-grained spatial cues and long-range temporal dynamics in payment recognition tasks.

	\item Design a prior-driven Spatial Difference Discriminator that adaptively learns anchor points and explicitly models local hand-to-anchor relative motion patterns to improve task-specific discriminability.

	\item We propose an integrated, deployment-oriented payment action recognition framework for onboard transit surveillance that leverages existing monitoring infrastructure and avoids extra hardware installation.

	\item Collect a real-world payment action dataset in collaboration with the local transit authority, featuring challenging onboard surveillance conditions.
	\end{itemize}

	%%%%%%%%%%%%%%%%%%%%%%%%%%%%%%%%%%%%%%%%%%%%%%%%%%%%%%%%%%%%%%%%%%
	\section{Related Work}

	\subsection{Transit Payment Analysis}
	Automatic transit payment analysis plays a critical role in intelligent fare collection systems and enables large-scale passenger behavior analytics and transit system monitoring \cite{bieler2022survey}. Traditional solutions mainly rely on manual inspection or physical barrier gates \cite{reddy2011measuring}, which are labor-intensive and limited in coverage. Recent studies have therefore explored vision-based automation for transit payment detection.

	Video-based approaches analyze surveillance footage to detect abnormal behaviors such as tailgating or barrier bypassing. However, appearance-driven methods are sensitive to adverse visual conditions, including low resolution, illumination variation, and occlusion \cite{huang2025transitreid}. To mitigate these issues, skeleton-based representations have been introduced to model human pose dynamics instead of raw pixels. Pose estimators such as OpenPose extract joint coordinates that can be further used for behavior modeling. Huang et al.~\cite{huang2022detection} leveraged skeleton sequences to identify fare evasion actions based on relative joint geometry, while \cite{huang2022time} modeled multi-person spatialtemporal interactions for tailgating detection. Nevertheless, these model-free approaches rely heavily on handcrafted motion statistics, limiting their representation power and robustness in highly noisy environments such as onboard bus videos.

	Multi-modal methods further enhance robustness by incorporating additional signals such as motion cues, audio streams, and RFID readings \cite{adanyin2024ai, wauyo2025towards}. However, these solutions depend on specialized sensing infrastructure that is not universally available in real deployments.

	Despite these advances, prior work primarily formulates the problem as binary fare evasion detection, overlooking the need for fine-grained understanding of payment behaviors. In modern open or barrier-free transit systems, diverse payment modalities such as cash payment, card swipe, QR code scanning, and tap-to-pay exhibit distinct motion patterns and operational implications, making their automatic recognition important for revenue auditing, passenger flow analytics, and system intelligence. To the best of our knowledge, fine-grained transit payment action recognition remains largely underexplored in the literature, motivating the proposed iPay framework.

	\subsection{Human Action Recognition}
	Human action recognition models spatiotemporal human motion from videos or pose signals. In surveillance settings, skeleton-based methods are appealing because they suppress appearance and background factors, offering a more privacy-friendly and robust alternative to RGB. Early approaches used handcrafted descriptors or RNNs such as hierarchical RNNs~\cite{du2015hierarchical} for temporal modeling, but they do not explicitly encode the joint topology needed for articulated-body dependency modeling.

	Graph Convolutional Networks (GCNs) address this limitation by representing a skeleton sequence as a spatiotemporal graph. 
	ST-GCN~\cite{yan2018spatial} established a strong baseline by applying graph convolutions over joints and temporal convolutions over frames, demonstrating effective motion modeling on large-scale benchmarks such as NTU RGB+D~\cite{shahroudy2016ntu}. 
	Subsequent work improved the expressiveness of the graph structure and feature representations. 
	For example, 2s-AGCN~\cite{shi2019two} introduced adaptive, learnable adjacency matrices and complementary joint--bone streams to enhance spatial modeling. 
	More recent variants further refine spatiotemporal message passing and channel modeling~\cite{chen2021channel,chi2022infogcn,zhou2024blockgcn}, and deformable graph mechanisms have been explored to better capture long-range dependencies and complex motion patterns under real-world noise~\cite{myung2024degcn}. 
	These advances make skeleton GCNs a natural backbone choice for motion-centric recognition in challenging settings.

	In parallel, RGB-based video understanding has progressed rapidly with strong 3D CNN and transformer backbones, such as SlowFast~\cite{feichtenhofer2019slowfast}, Video Swin~\cite{liu2022video}, and masked video pretraining frameworks like VideoMAE~\cite{tong2022videomae}. 
	While these models achieve impressive performance on consumer-grade benchmarks, their effectiveness can degrade in onboard surveillance scenarios where the target action occupies a small region, visual quality is low, and background distractions are severe. 
	This gap is particularly pronounced for fine-grained actions dominated by localized hand motion and subtle hand-object interactions, where global scene cues provide limited discrimination.

	These observations motivate multimodal learning that leverages the complementary strengths of RGB and skeleton cues. Existing multimodal frameworks often adopt late fusion or loosely coupled experts~\cite{bruce2022mmnet}, which is suboptimal for fine-grained, domain-specific behaviors. In onboard payment recognition, the discriminative evidence is both temporal (articulated motion) and spatial (local hand-terminal interaction), yet identity-dependent appearance cues should be minimized. 
	Therefore, iPay builds on an effective video backbone and a modern skeleton GCN, and further introduces feature-level cross-modal coupling and a explixit adatively-learned motion prior to better capture localized payment dynamics.

	%%%%%%%%%%%%%%%%%%%%%%%%%%%%%%%%%%%%%%%%%%%%%%%%%%%%%%%%%%%%%%%%%%
	\section{Methodology}
	\subsection{Preliminaries}
	In this section, we introduce the skeleton data extraction and the spatiotemporal graph formulation used by the skeleton stream. We follow the common ST-GCN \cite{yan2018spatial} convention where each skeleton sequence is represented as a graph whose nodes are body joints and edges capture both physical connections and temporal correspondences.

	\paragraph{Skeleton Extraction}
	A number of benchmark datasets have been widely used for human action recognition, such as NTU RGB+D~\cite{shahroudy2016ntu}, NTU RGB+D 120~\cite{liu2019ntu}, and NW-UCLA~\cite{wang2014cross}. However, due to the domain gap, these datasets cannot be directly used to train our payment-action detection model. We therefore collect a dedicated dataset (see Section~Experiments for details). To extract skeletons, we adopt SAM3~\cite{carion2025sam} with the \textit{sam-3d-body} model to segment human bodies and estimate 3D skeletons from videos. Each sample is finally represented as a skeleton tensor 
	\begin{equation}
	\mathbf{X} \in \mathbb{R}^{M \times C_{in} \times T \times V}
	\end{equation}
	where $C_{in}=3$ denotes the number of input channels, $T$ is the number of frames, $V$ is the number of joints, and $M$ is the maximum number of persons per frame. 

	To suppress translation and scale variation across frames, we center each skeleton at its pelvis and normalize it by the hip distance. Specifically, let $\mathbf{p}_{t,v}\in\mathbb{R}^{3}$ denote the 3D coordinate of joint $v$ at frame $t$, and let $h_L,h_R$ denote the left and right hip joints. We define the per-frame scale factor $s_t=\lVert \mathbf{p}_{t,h_L}-\mathbf{p}_{t,h_R}\rVert_2$ and compute pelvis-aligned, scale-normalized coordinates 
	\begin{equation}
	\tilde{\mathbf{p}}_{t,v}=(\mathbf{p}_{t,v}-\mathbf{p}_{t,p})/s_t
	\end{equation}
	where $\mathbf{p}_{t,p} = (\mathbf{p}_{t,h_L}+\mathbf{p}_{t,h_R})/2$ denotes the pelvis point. This normalization reduces inter-subject scale differences and removes global translation, thereby emphasizing subtle inter-joint motion.

	\paragraph{Spatiotemporal graph Convolution}
	For mini-batch training with batch size $N$, the input skeleton tensor is
	\begin{equation}
	\mathbf{X_{in}} \in \mathbb{R}^{N \times M \times C_{in} \times T \times V}
	\end{equation}

	We then define a spatiotemporal graph $G=(\mathcal{V},\mathcal{E})$ with node set
	\begin{equation}
	\mathcal{V}=\{(t,v)\mid t \in \{1,\dots,T\},\ v \in \{1,\dots,V\}\}
	\end{equation}
	and edge set $\mathcal{E}=\mathcal{E}_S \cup \mathcal{E}_T$. The spatial edges $\mathcal{E}_S$ connect anatomically linked joints within the same frame, while the temporal edges $\mathcal{E}_T$ connect the same joint across adjacent frames. The corresponding adjacency matrix is denoted by $\mathbf{A} \in \mathbb{R}^{(TV)\times(TV)}$, and $\hat{\mathbf{A}}$ denotes a normalized adjacency used for message passing.

	Let $\mathbf{X}^{(l)} \in \mathbb{R}^{(TV)\times C_l}$ be the node feature matrix at layer $l$, where $C_l$ is the feature dimension. A standard graph convolution updates node features by aggregating neighbors:
	\begin{equation}
	\mathbf{X}^{(l+1)} = \sigma\!\left(\hat{\mathbf{A}}\,\mathbf{X}^{(l)}\mathbf{W}^{(l)}\right)
	\end{equation}
	where $\mathbf{W}^{(l)} \in \mathbb{R}^{C_l \times C_{l+1}}$ is a learnable weight matrix, and $\sigma(\cdot)$ denotes the ReLU activation function.

	\subsection{Skeleton Feature Stream}
	Based on the above spatiotemporal graph formulation, we build a skeleton feature stream to capture motion dynamics from body joints. Specifically, the stream takes the normalized skeleton sequence as input and applies a stack of spatiotemporal graph convolutional blocks to learn discriminative motion representations. We choose DeGCN~\cite{myung2024degcn} as our backbone, since its deformable convolution module can effectively capture spatiotemporal contextual relationships. The resulting features $\mathbf{F_s} \in \mathbb{R}^{N \times M \times C_s \times T_s \times V_s}$ are globally pooled over the temporal and joint dimensions and fed into a lightweight linear head and a softmax to produce the skeleton logits $\mathbf{L_S}$. This skeleton branch serves as a motion expert and provides complementary temporal cues for subsequent cross-modal fusion.

	\subsection{RGB Feature Stream}
	Although the skeleton stream captures fine-grained motion, it tends to overlook subtle local relative motions, while per-frame alignment and normalization inevitably discard absolute spatial cues. Thus, we incorporate an RGB stream to recover region-level semantics across space and time, such as interaction location and the handled object.

\begin{figure}[t]
		\centering
		\includegraphics[width=\columnwidth]{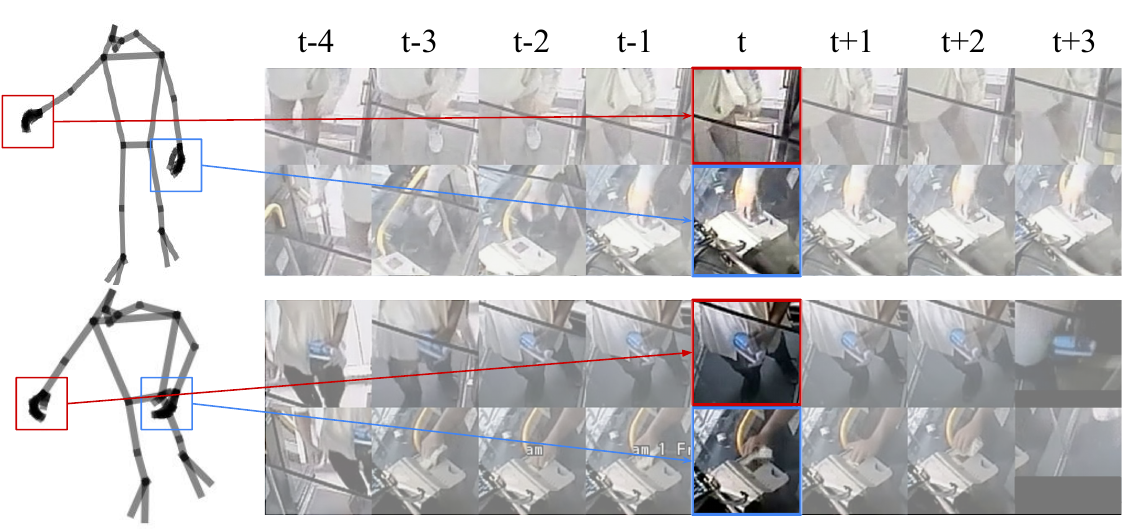}
		\caption{Construction of the spatiotemporal ROI (ST-ROI). Action-relevant regions are cropped from each frame and concatenated in temporal order to form a single image for the RGB stream.}
		\label{fig:st-roi}
	\end{figure}

	Since the input comprises video frames, many approaches employ LSTMs, 3D convolutions, or Vision Transformers to model temporal dynamics \cite{carreira2017quo, zhao2022real, pang2025context}. However, these solutions often incur substantial computational cost and require more complex architectures. Since our primary goal is to capture fine-grained spatial cues, we follow MMNet \cite{bruce2022mmnet} and construct a spatiotemporal region of interest (ST-ROI) for each video sample, as illustrated in Fig.~\ref{fig:st-roi}. The cropped action-relevant regions from each frame preserve fine-grained spatial semantics while encoding motion as a contiguous pattern.

	\paragraph{ST-ROI Construction}
	Let a video clip be uniformly sampled into $L$ frames $\{I_t\}_{t=1}^{L}$. For each frame, we first estimate 2D body-joint locations and convert them into a compact region representation by selecting a subset of informative joints and computing bounding boxes centered on these joints. To ensure a consistent input size for the RGB backbone, we resize the cropped region to a fixed resolution and clip the box coordinates to the image boundary when necessary. We then concatenate all boxes vertically to obtain per-frame ROIs $\{b_t\}_{t=1}^{L}$ that guide the RGB stream to focus on action-relevant regions while suppressing background clutter. To incorporate temporal context with minimal overhead, we concatenate the ROIs from all $L$ frames in temporal order along the width dimension to form a single ST-ROI image $\mathbf{R} \in \mathbb{R}^{H \times WL \times 3}$, which preserves the temporal structure of the motion and enables standard 2D CNN backbones to model temporal evolution without recurrent units or 3D convolutions. The constructed ST-ROI is finally fed into the RGB feature stream to extract region-level spatiotemporal semantics.

	\paragraph{RGB Feature Head}
	In Fig.~\ref{fig:framework}, the RGB feature head takes the constructed ST-ROI image as input and uses a ResNet-18 \cite{he2016deep} backbone to extract a compact feature representation. The resluting RGB embeddings $\mathbf{F_r} \in \mathbb{R}^{N \times C_r \times H_r \times W_r}$ is then projected through a small linear head and a softmax to produce the action logits $\mathbf{L_R}$. This RGB branch serves as a context expert and provides complementary spatial cues for subsequent cross-modal fusion.

	\subsection{Dual-Attention Feature Stream}
	As discussed above, the RGB stream excels at capturing spatial details, whereas the skeleton stream is more effective at modeling temporal dynamics. To leverage the complementary strengths of these two modalities, prior work typically adopts post-fusion, where each modality independently produces action logits and the final prediction is obtained by a weighted average of their scores. Although several studies explore using skeleton-expert features to guide the RGB stream~\cite{bruce2022mmnet,cho2025body}, they do not enable the two modalities to be jointly optimized and fully integrated at the feature level. We therefore propose a Dual-Attention Feature Fusion module that couples cross-attention with gated fusion, allowing the model to learn fine-grained spatiotemporal cues from both modalities in a unified manner.

	The network architecture is illustrated in Fig.~\ref{fig:framework}. We first extract the skeleton feature $\mathbf{F_s}$ using the GCN branch and the RGB feature $\mathbf{F_r}$ using the RGB backbone. The two features are then fed into a bi-directional cross-attention module: in one direction, $\mathbf{F_s}$ serves as the query while $\mathbf{F_r}$ provides the keys and values; in the other direction, the roles are reversed. This design allows the two modalities to guide each other, enabling the skeleton stream to acquire richer spatial context from RGB cues and the RGB stream to capture temporal dependencies from skeleton dynamics.

	The two attention branches produce their respective embeddings $[\mathbf{F_s^{\mathrm{fused}}}, \mathbf{F_r^{\mathrm{fused}}}]$, which are concatenated and passed through a gated linear layer followed by a softmax to obtain the gate weights $[G_S, G_R]$. We then reweight the fused features and compute the final fused logits as
	\begin{equation}
	\mathbf{L_F}=\mathrm{Softmax}(G_S \cdot \mathbf{F_s^{\mathrm{fused}}} + G_R \cdot \mathbf{F_r^{\mathrm{fused}}})
	\end{equation}

	\subsection{Spatial Difference Discriminator Stream}
\begin{figure}[t]
		\centering
		\includegraphics[width=\columnwidth]{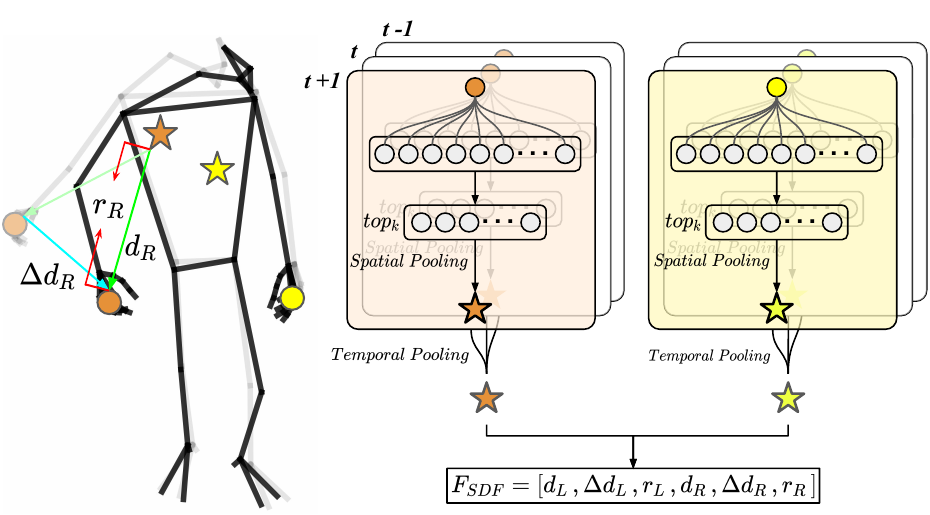}
		\caption{Overview of the proposed Spatial Difference Discriminator (SDD), which learns adaptive spatial anchors and encodes hand-centric motion dynamics for payment behavior recognition. Orange/yellow circles: right/left hand representative joints; orange/yellow stars: adaptively learned right/left anchor points.}
		\label{fig:SDD}
	\end{figure}

	The previous three streams mainly map the whole skeleton graph into a feature space in an implicit manner, lacking task-specific inductive bias. To address this limitation, we design a Spatial Difference Discriminator (SDD) as shown in Fig. \ref{fig:SDD}, which explicitly introduces an additional geometric cue to enhance the model's adaptability for the payment detection task. For payment behaviors, the motion center typically lies in the relative movement between the hand and the payment box. For example, the key difference between \textit{swipe} and \textit{tap} is that the former exhibits a clear lateral sliding motion with respect to the box, whereas the latter mainly involves a short contact-and-hold pattern. A straightforward way to explicitly model this prior is to construct vectors from hand joints to an anchor point on the payment box. However, manually annotating the payment box for every test scenario is impractical for real-world deployment. An alternative is to detect the box using a vision detector and sample anchor points accordingly, but in in-vehicle surveillance videos the detection is often unstable due to noise, occlusion, motion blur, and glass reflections.

	These observations motivate us to ask: can the model automatically learn the most discriminative anchor location? To this end, we propose the SDD module, which adaptively aggregates full-body joint information via attention pooling to learn two spatial anchors (for the left and right hands), thereby constructing robust hand-centric motion representations. 

	\paragraph{Adaptive Anchor Learning}

	Given the spatial features from the GCN backbone, we first perform joint-wise attention pooling. The temporally averaged joint features are fed into a lightweight MLP to produce attention weights:

	\begin{equation}
	w_v = \mathrm{Softmax}(\mathrm{MLP}(\mathbf{f}_v)), v \in V
	\end{equation}

	Using the top-$K$ weights, after normalization, we compute a soft spatial anchor as follows:

	\begin{equation}
	\mathbf{p}_A(t) = \sum_{v=1}^{K} w_v \, \mathbf{X}(t, v), \sum w_v = 1
	\end{equation}

	For temporal stability, the anchor is further averaged over time to obtain global anchor points $\mathbf{p}_{AL}$ and $\mathbf{p}_{AR}$.

	\paragraph{Hand-Centric Spatial Encoding}

	Let $\mathbf{x}_L(t)$ and $\mathbf{x}_R(t)$ denote the coordinates of the left and right hand joints. We compute hand-to-anchor displacement vectors:
	\begin{equation}
	\mathbf{d}_L(t) = \mathbf{x}_L(t) - \mathbf{p}_{AL},\quad
	\mathbf{d}_R(t) = \mathbf{x}_R(t) - \mathbf{p}_{AR}
	\end{equation}

	To capture motion dynamics, we further compute first-order temporal differences:

	\begin{equation}
	\Delta \mathbf{d}(t) = \mathbf{d}(t) - \mathbf{d}(t-1)
	\end{equation}

	and the corresponding Euclidean magnitudes:

	\begin{equation}
	r(t) = \|\mathbf{d}(t)\|_2
	\end{equation}

	The final per-frame spatial discriminative feature is constructed as

	\begin{equation}
	\mathbf{F}_{SDF}(t) =
	[\mathbf{d}_L,\,
	\Delta\mathbf{d}_L,\,
	r_L,\,
	\mathbf{d}_R,\,
	\Delta\mathbf{d}_R,\,
	r_R]
	\end{equation}

	resulting in a 14-dimensional hand-centric descriptor.

	\paragraph{Temporal Modeling}

	The feature sequence is fed into a lightweight two-layer Temporal Convolution Network (TCN), followed by global pooling and a fully connected classifier to produce the final prediction $\mathbf{L_{SDD}}$. By explicitly encoding adaptive spatial anchors and hand-relative motion dynamics, SDD provides a strong task-oriented inductive bias for fine-grained payment behavior recognition, while remaining lightweight and fully differentiable.

	%%%%%%%%%%%%%%%%%%%%%%%%%%%%%%%%%%%%%%%%%%%%%%%%%%%%%%%%%%%%%%%%%%
	\begin{table*}[t]
		\centering
		\caption{Summary of the Collected Transit Payment-action Dataset.}
		\label{tab:dataset}
		\begin{tabular}{ccccc}
			\toprule
			Label & Payment Method & Description & Number Recorded & Percentage \\ \midrule
			0 & QR Code & Scan QR code on mobile device & 124 & 21.8\% \\ 
			1 & Cash & Insert bills and/or coins & 97 & 17.0\% \\ 
			2 & Evade & No payment made & 37 & 6.5\% \\ 
			3 & Swipe & Swipe payment card & 88 & 15.5\% \\ 
			4 & Tap & Tap payment card & 145 & 25.5\% \\ 
			5 & Others & Difficult to determine label & 78 & 13.7\% \\ \midrule
			- & Total & All payment methods & 569 & 100\% \\ \bottomrule
		\end{tabular}
	\end{table*}

	\section{Experiments}
	\subsection{Dataset}
	Since existing human action recognition datasets focus on general action recognition~\cite{shahroudy2016ntu, liu2019ntu, wang2014cross}, there is a substantial domain gap between these datasets and our public transportation payment-action setting. Therefore, we collect a dedicated dataset from buses in real operation. Specifically, we gather over 55 hours of real surveillance video from 16 bus-route recordings and manually annotate passengers' payment behaviors. Skeleton data are extracted using SAM3, with 70 joints following the MHR70 format, which provides fine-grained hand and finger keypoints, enabling precise modeling of subtle hand-object interactions critical for payment behavior recognition. For each frame, only one passenger is retained. The dataset is summarized in Table~\ref{tab:dataset}. During training, the train/test split is 0.7/0.3.

	\subsection{Implementation Details}
	Experiments are conducted using the PyTorch framework on four A6000 GPUs. DeGCN~\cite{myung2024degcn} is selected as the skeleton backbone and ResNet-18~\cite{he2016deep} as the RGB backbone. The dual-stream setting in~\cite{myung2024degcn} is retrained, and with one stream for modality fusion. The number of sampled frames $L$ is set to 8, and the informative joints \texttt{left\_middle\_finger3} and \texttt{right\_middle\_finger3} are chosen as defined in the MHR70 format. The number of aggregated neighbors $K$ is set to 8. The final prediction is obtained by summing the logits from the four streams and selecting the class with the highest score. The model is trained for 80 epochs using SGD with a hybrid learning-rate schedule, where the learning rate is first linearly warmed up to 0.06, then decayed via cosine annealing, and finally switched to step decay. For stable training, each sample is uniformly sampled to 100 frames. The data are randomly shifted and rotated for data augmentation. In addition, due to the unclear payment patterns of the ``Others'' category, we exclude it from both training and testing.

	For comparisons with previous methods, we follow the same training settings to ensure fairness, but adjust the learning rate to 1e-2 for ST-GCN \cite{yan2018spatial} and to 1e-3 for InfoGCN \cite{chi2022infogcn} to achieve the best performance. We also manually design the multi-scale graph following \cite{chi2022infogcn, zhou2024blockgcn} and the hypergraph following \cite{zhu2022multilevel}.

	\subsection{Comparison With State-of-the-Art}
	In this subsection, we compare the proposed network (iPay) with previous state-of-the-art methods on the collected payment recognition dataset. It is worth noting that most existing methods employ two or four ensembles with post-logit fusion, whereas our method is end-to-end, leading to more stable training and reduced inference burden. Table \ref{tab:comparison} shows that our iPay (2-stream) outperforms existing approaches in payment recognition. The iPay 1-stream denotes using only a single GCN stream from the baseline, while the 2-stream version represents our full model, which achieves the best performance at the cost of increased computational overhead. Compared with the pioneering GCN-based action recognition model ST-GCN \cite{yan2018spatial}, our method improves the average accuracy by 23.02\%. Compared with the baseline DeGCN \cite{myung2024degcn}, the improvement is 10.07\%. Interestingly, although DeGCN is currently SOTA on some benchmarks, CTR-GCN \cite{chen2021channel} performs better on our dataset, suggesting that generally strong methods may not always be optimal for task-specific scenarios. In terms of model size, due to the introduction of the RGB backbone, iPay has a larger number of parameters. However, its FLOPs (floating-point operations) are lower than all models except InfoGCN \cite{chi2022infogcn}, achieving competitive computational efficiency with only 5.07G/6.34G FLOPs.

	For the per-class accuracy, most existing approaches exhibit a noticeable class imbalance issue, particularly on the Swipe and QR Code categories. In contrast, iPay demonstrates a more balanced performance across different payment types, which validates the effectiveness of the proposed network designs tailored for payment behavior.

	\begin{table*}[t]
		\centering
		\caption{(Per-class) Accuracy (\%), FLOPs, and Parameter Comparison with State-of-the-art Methods on the Proposed Payment Recognition Dataset. $\dagger$ and $\dagger\dagger$ Indicate Fusion Results with Two and Four Ensembles, Respectively. The Best and Second-best Results are Highlighted in \textbf{Bold} and \underline{Underline}, Respectively.}
		\label{tab:comparison}
		\begin{tabular}{lc|cccccc|cc}
			\toprule
			Method & Source & QR Code & Cash & Evade & Swipe & Tap & Avg. & FLOPs & Params. \\ \midrule
			ST-GCN \cite{yan2018spatial} & AAAI18 & 64.86 & 73.91 & \textbf{100} & 38.46 & 52.38 & 60.43 & 7.62G & 3.07M \\
			CTR-GCN$\dagger\dagger$ \cite{chen2021channel} & ICCV21 & 64.86 & \underline{86.96} & 81.82 & 69.23 & \underline{83.33} & 76.26 & 11.04G & 5.64M \\
			InfoGCN$\dagger$ \cite{chi2022infogcn} & CVPR22 & 54.05 & \textbf{91.30} & 72.73 & 53.85 & 71.43 & 66.91 & 4.66G & 3.06M \\
			ML-STGNet$\dagger$  \cite{zhu2022multilevel} & TIP22 & \underline{72.97} & 82.61 & \underline{90.91} & 65.38 & 80.95 & 76.98 & 6.12G & 5.64M \\
			DeGCN$\dagger\dagger$ (baseline) \cite{myung2024degcn} & TIP24 & 70.27 & 82.61 & \textbf{100} & 46.15 & 80.95 & 73.38 & 9.80G & 5.24M \\
			BlockGCN$\dagger\dagger$  \cite{zhou2024blockgcn} & CVPR24 & 54.05 & 82.61 & \underline{90.91} & 57.69 & 78.57 & 69.78 & 9.88G & 4.84M \\
			\midrule
			\textbf{iPay$\dagger\dagger$, 1-stream} & Ours & \textbf{75.68} & 82.61 & 81.82 & \underline{76.92} & 76.19 & \underline{77.70} & 5.07G & 12.88M \\
			\textbf{iPay$\dagger\dagger$, 2-stream}  & Ours & \underline{72.97} & 82.61 & \underline{90.91} & \textbf{88.46} & \textbf{88.10} & \textbf{83.45} & 6.34G & 13.55M \\
			\bottomrule
		\end{tabular}
	\end{table*}

	% \subsection{Qualitative Analysis}

	\subsection{Ablation Studies}
	\paragraph{Contribution of Different Components}
	Table \ref{tab:components} reports the performance under different modality compositions. We first evaluate the baseline with different input modalities, where \textit{4-ensemble} denotes late logit fusion over all modalities as in \cite{myung2024degcn}, which performs substantially better than any single-modality setting. We then examine our RGB-only branch, which achieves only 53.96\% accuracy, largely due to the limited visual quality of onboard surveillance footage. After jointly training with Joint and RGB experts (\textit{2-ensemble}), the accuracy increases to 80.58\%, outperforming the Joint-only baseline (64.75\%) by 15.83\%, confirming the clear benefit of multimodal learning for this task. Introducing the proposed dual-attention fusion (\textit{3-ensemble}) further improves performance to 82.01\%, indicating that deep, feature-level coupling enables the two modalities to compensate for each other's weaknesses. Finally, adding the Spatial Difference Discriminator (SDD) yields our full model with 83.45\% accuracy, demonstrating that explicitly modeling local hand-to-anchor relative motion enhances sensitivity to such payment actions dominated by strong localized dynamics. Notably, our model uses only the joint modality from the skeleton representation; hereafter, “skeleton” refers to joints unless stated otherwise.

	Moreover, our full model requires fewer FLOPs (6.34G) than the baseline (9.80G), indicating that it achieves higher accuracy while maintaining fast inference, which is well-suited for edge deployment in transportation systems. The number of parameters increases due to the inclusion of the ResNet backbone (11.18M).

	\begin{table}[t]
		\centering
		\caption{Comparison with Multi-modal Ensemble. *-ensemble Denotes Utilizing * Streams.}
		\label{tab:components}
		\begin{tabular}{l|ccc}
			\toprule
			Method & Acc. (\%) & FLOPs & Params. \\ \midrule
			Baseline (only Joint)         & 64.75 & 2.45G & 1.31M \\
			Baseline (only Bone)          & 68.35 & 2.45G & 1.31M \\
			Baseline (only Velocity)      & 55.40 & 2.45G & 1.31M \\
			Baseline (4-ensemble)         & 73.38 & 9.80G & 5.24M \\ 
			\midrule
			RGB                           & 53.96 & 3.72G & 11.18M \\
			Joint+RGB (2-ensemble)        & 80.58 & 6.25G & 12.51M \\
			Joint+RGB (3-ensemble)        & 82.01 & 6.32G & 13.31M \\
			\midrule
			Joint+RGB+SDD (4-ensemble)    & 83.45 & 6.34G & 13.55M \\
			\bottomrule
		\end{tabular}
	\end{table}

	\paragraph{Dual-Attention vs. Single-Attention}
	We evaluate different attention configurations as reported in Table \ref{tab:attention}. With a single-attention design using RGB as the query, the accuracy drops to 30.22\%, consistent with the weak performance of the RGB-only stream in Table \ref{tab:components}. In contrast, using the skeleton modality as the query achieves 81.29\% accuracy, and the proposed dual-attention setting further improves it to 83.45\%. This indicates that bidirectional cross-modal interaction allows each modality to benefit from the complementary strengths of the other.

	\begin{table}[t]
		\centering
		\caption{Comparison with Different Attention Settings. RGB/Ske Denotes Using that Modality As Query.}
		\label{tab:attention}
		\begin{tabular}{l|ccc}
			\toprule
			Method & Acc. (\%) & FLOPs & Params. \\
			\midrule
			Single-Attn-RGB (4-ensemble)      & 30.22 & 6.30G & 13.15M \\
			Single-Attn-Ske (4-ensemble)      & 81.29 & 6.30G & 13.15M \\
			Dual-Attn (4-ensemble)            & 83.45 & 6.34G & 13.55M \\
			\bottomrule
		\end{tabular}
	\end{table}

	\paragraph{Data Pre-processing}
	We compare different skeleton input variants and alignment strategies, as shown in Table~\ref{tab:processing}. A natural intuition for payment action recognition is that the behavior mainly depends on the upper body, especially the arms and hands, while the lower body contributes little. To test this, we remove the 10 lower-body joints. However, the resulting performance is worse than using the full-body joints (80.58\% vs.\ 83.45\%). An inspection of the data suggests that lower-body cues correlate with subtle differences in stance duration and body orientation across payment types. For example, cash payments involve longer standing and a more frontal posture, while swipe and tap are typically brief and more side-facing. This trend also appears at class-level results: with upper-body joints only, QR and Cash achieve 75.68\% and 95.65\% accuracy, higher than the full-body setting, while Swipe and Tap drop to 73.08\% and 78.57\%. Overall, using the full-body skeleton provides a more general representation that better balances performance across classes.

	In preprocessing, we align each frame by the pelvis to suppress global translation. With alignment enabled, accuracy improves by 0.72\%. Although modest, this gain supports our hypothesis that payment behaviors are dominated by localized relative motion patterns, and it is consistent with the design motivation of our hand-centric modeling components.

	\begin{table}[t]
		\centering
		\caption{Comparison with Different Data Preprocessing. w/o Alignment Denotes Aligning by the First Frame of Each Clip.}
		\label{tab:processing}
		\begin{tabular}{l|ccc}
			\toprule
			Method & Acc. (\%) & FLOPs & Params. \\
			\midrule
			Upper Body (4-ensemble)      & 80.58 & 5.95G & 13.55M \\
			Whole Body (4-ensemble)      & 83.45 & 6.34G & 13.55M \\
			\midrule
			w/o Alignment (4-ensemble)   & 82.73 & 6.34G & 13.55M \\
			\bottomrule
		\end{tabular}
	\end{table}

	%%%%%%%%%%%%%%%%%%%%%%%%%%%%%%%%%%%%%%%%%%%%%%%%%%%%%%%%%%%%%%%%%%
	\section{Conclusions}
	This paper presents iPay, an integrated payment action recognition framework tailored for onboard transit surveillance systems. iPay addresses real-world transit payment recognition challenges at both the system and algorithm levels by combining privacy-aware local RGB evidence with robust skeletal motion cues. To bridge the modality gap in fine-grained payment behaviors, we propose a multimodal mixture-of-experts architecture with four tightly coupled streams. We also create a real-world payment-action dataset from over 55 hours of onboard surveillance footage. Experiments show that iPay outperforms strong skeleton-based baselines and recent state-of-the-art methods on this challenging dataset, while maintaining competitive computational efficiency for edge deployment in transportation systems. Future work will prioritize real-world deployment and improve robustness to severe data degradation (e.g., occlusion, reflections, and missing observations).
	%%%%%%%%%%%%%%%%%%%%%%%%%%%%%%%%%%%%%%%%%%%%%%%%%%%%%%%%%%%%%%%%%%
	\section*{ACKNOWLEDGMENTS}
	% Replace with acknowledgments or remove if none
	The authors would like to thank the NVIDIA Academic Grant Program for its support. The authors also gratefully acknowledge Bill Ray and Calvin Young from the Capital District Transportation Authority for their valuable support.
	
	%%%%%%%%%%%%%%%%%%%%%%%%%%%%%%%%%%%%%%%%%%%%%%%%%%%%%%%%%%%%%%%%%%
	%\addtolength{\textheight}{-12cm}
	%\vspace{10mm}
	\bibliographystyle{IEEEtran}
	% Your .bib file here
	\bibliography{root} 
	
\end{document}